# Rule-Based Classification of Hyperspectral Imaging Data

**Songuel Polat [1,2], Alain Tremeau [2] and Frank Boochs [1]**

[1] i3mainz, Institute for Spatial Information and Surveying Technology, Mainz University of Applied Sciences, Lucy-Hillebrand-Str. 2, 55128 Mainz, Germany; songuel.polat@hs-mainz.de; frank.boochs@hs-mainz.de

[2] Hubert Curien Laboratory, University Jean Monnet, University of Lyon, 18 Rue Professeur Benoît Lauras, 42100 Saint-Etienne, France; alain.tremeau@univ-st-etienne.fr

**Abstract:** Due to its high spatial and spectral information content, hyperspectral imaging opens up new possibilities for a better understanding of data and scenes in a wide variety of applications. An essential part of this process of understanding is the classification part. In this article we present a general classification approach based on the shape of spectral signatures. In contrast to classical classification approaches (e.g. SVM, KNN), not only reflectance values are considered, but also parameters such as curvature points, curvature values, and the curvature behavior of spectral signatures are used to develop shape-describing rules in order to use them for classification by a rule-based procedure using IF-THEN queries. The flexibility and efficiency of the methodology is demonstrated using datasets from two different application fields and leads to convincing results with good performance.

## 1. Introduction

Optical technology developments are extending the possibilities to better understand the world and its resources. Starting with images consisting of three color channels covering the visual electromagnetic spectrum, the developments since the late 1960s opened up the possibility of using spectral properties for identification of materials by using multispectral images with tens of channels. Especially with the developments in the last two decades, another enormous step forward was made and the low spectral resolution of multispectral images was overcome. With several hundred narrow channels, hyperspectral imaging (HSI) opens up completely new possibilities for analysis in a wide variety of application fields [1]. As examples, [2–4] use HSI to maintain and increase crop yields in precision agriculture. The evaluation of food quality and safety through the use of HSI is part of the research of [5–7]. Other applications of HSI can be found in medicine to diagnose diseases or to monitor wound healing [8–10], in the art market to verify the authenticity of artworks [11–13] and in forensics to analyze crime scenes [14–16].

The key technology behind those applications is the HSI with detailed spectral and spatial information, which makes the HSI a powerful information source for advanced classification methods like k-nearest neighbor, support vector machines, random forests, neural networks and deep learning approaches. A comparison of these mentioned classification methods is conducted in [17] and shows, that there is no classifier that consistently provides the best performance and that the quality of the classification result mainly depends on factors such as the availability of training samples, processing requirements, tuning parameters and speed of the algorithm.

Another aspect to be considered in the context of mentioned classifiers is the Curse of Dimensionality. If high dimensional HSI are directly used as input, the classification accuracy decreases, while the computational effort of the model tends to increase exponentially. To avoid this problem, a dimensionality reduction is essential [18]. The reduction of dimensions implies that algorithms automatically have to extract a set of characteristic spectral values, which have to represent the deciding features of the objects in question, from the entire course of a spectral signature. The amount of existing studies (e.g. [19–24]) proves that there is no existing satisfactory, robust and reliable methodology and that this topic is one of the main open issues in spectral imaging. As a consequence, this recommended step leads to loss of information in the spectral space, because the selected bands do not give an accurate description of the original spectral signature.

Other classification approaches based on indices [25–27] also arbitrarily select a subset of spectral values. One of the most popular is the Normalized Difference Vegetation Index (NDVI) from the field of remote sensing [28], which combines only a few bands of NIR and Red wavelengths. By limiting to a small number of spectral bands in a restricted spectral range, broad classifications such as separation between plants, water or urbanized areas can be made, but finer separation (e.g. between plant species) would be difficult due to the reduced number of spectral bands and contradicts the use of HSI, which offers the enormous advantage of high-resolution spectral information.

Both, band selection methods and indices consider only a small set of characteristic values, while the shape of the spectra is not taken into account. A consideration of the shape of spectral curves as important information source can be found in [29–31].

The work of [30] consists of a classification approach that fully utilizes the shape of spectral curves by using a code to parameterize the spectral curve shape. The research of [31] deals with the definition of analysis rules based on spectral features like band position, band depth, band width and band asymmetry. These key parameters are used to describe the absorption features of spectra. Disadvantage of both works are the time required to develop the description parameters of the curve shapes. Both approaches use tables to store the parameters and a subsequent matching process between the parameters of the reference data and the spectra that needs to be classified, which also makes the classification process a time consuming task.

The advantage of considering shapes, especially in combination with high resolution spectral data, is clear: spectra express the mixed reflectivity of the elements that make up an object (e.g. molecules, pigments, cell structure, water content), which is why each individual component has only a proportional influence on a spectrum. Changes in the composition of the elements then mainly change the mixture of all spectral contributions, resulting in local or regional variations and changing the shape of a spectrum.

In this article we follow the shape-based works of [29–31] and propose a new Rule-based classification method using shape-based properties of spectral curves like curvature points, curvature values, curvature direction and spectral values.

Particular attention should be paid to the following points:

- The formulation of the rules should not become complex.
- The classification process should not be time consuming.

Keeping these points in mind, an approach was developed that allows to establish rules, using any kind of logical elements, in a straightforward manner. The establishment of rules implies existing knowledge. A prior analysis of the data and the acquisition of knowledge enable a better understanding of the data and allow to structure and simplify a problem. Research papers from the field of remote sensing, that use the advantage of knowledge can be found in [32–37]. The experimental part for the evaluation of the method was performed on two different datasets. A detailed description of the method and the used datasets are part of section 2.

## 2. Materials and Methods

### 2.1 Spectral Data Set Acquisition

The hyperspectral systems used in this work are two pushbroom cameras from Specim Ltd. (Oulu, Finland). The Specim FX10 captures the spectral signature from 400 nm to 1000 nm (233 bands), while the Specim FX17 captures the spectral signature from 900 nm to 1700 nm (229 bands). All spectral images acquired by these cameras are radiometrically normalized by using dark reference images for dark-current (closed shutter) and a white reference image to reduce the influence of the intensity variability. For the white calibration a 99% reflectance tile was used.

As shown in Figure **1**, a set of different HSI consisting of two different object types are used to demonstrate the proposed method.

The first dataset was captured with the Specim FX17 and shows an image of seven classes of different plastic types. It has a resolution of 661 x 500 pixels and 229 bands with a spectral range from 900 nm to 1700 nm. The second scene was captured with the Specim FX10 and the Specim FX17 and shows ten classes of different plant types. The HSI of both cameras were combined and have a resolution of 1220 x 640 pixels and 462 bands covering the spectral range from 400 nm to 1700 nm.

The HSI with different plastic types is used to demonstrate the basic functionality of the approach and gets more attention due to existing ground truth[1]. As part of a waste sorting application, the demonstration also covers experiments with real waste consisting of plastics and electronic waste like printed circuit boards (PCB). The plant-based dataset, on the other hand, is used to demonstrate the potential and flexibility of the approach regarding different kind of applications.

---

[1] The datasets can be found on https://doi.org/10.5281/zenodo.5068201.

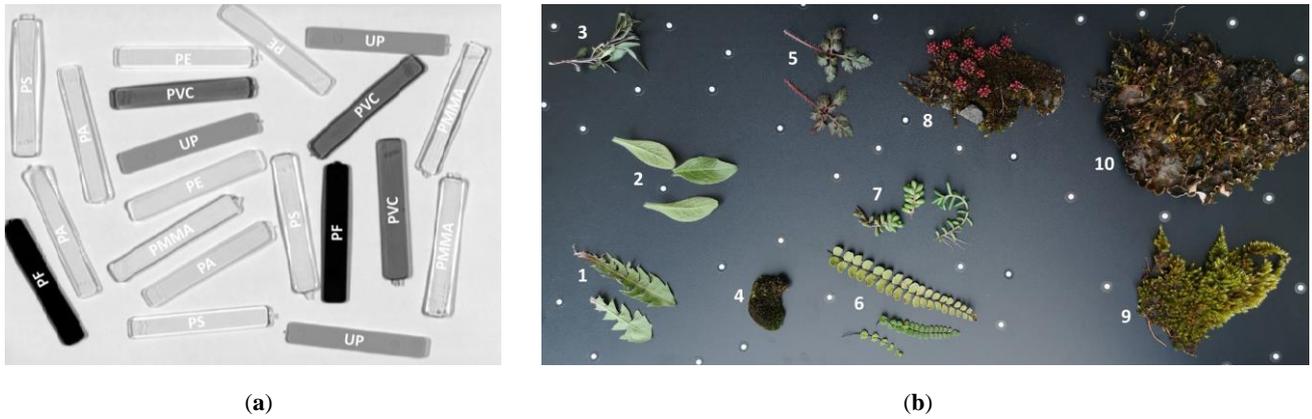

(a) (b)

**Figure 1.** Images of the captured scenes (a) Seven classes of different plastics[2] consisting of phenolic resin (PF), polyamide (PA), polyvinylchloride (PVC), polyethylene (PE), polymethylmethacrylate (PMMA), polyester resins (UP) and polystyrene (PS); (b) Ten classes of different vegetation types consisting of taraxacum (1), inula conyza (2), campanula rotundifolia (3), moss (4), geranium robertianum (5), asplenium trichomanes (6), green sedum (7), moss & red sedum (8), moss (9), moss & lichen (10).

## 2.2 Methodology

Spectral signatures offer the ability to distinguish between materials and are the result of reflected light from the surface, which is captured within a broad electromagnetic spectrum. The reason for different signatures is a combination of molecules in materials and the morphological structure. Different molecules result in different spectral signatures. The morphological structure, on the other hand, is important because of the resulting light path that is created by reflection, absorption, transmission and deflection from the different components of an object. That is why plants show different spectra when the cell structure changes, e.g. due to stress or aging. Both, differences in molecules and structure, result in different shaped spectral curves. The idea of our work is to describe the spectral curve shape by a combination of spectral values and shape-based parameters, to use this knowledge for the formation of rules.

Changes in the material composition of objects lead to local or regional changes in the course of spectral signatures. These changes inevitably lead to changes in the curvature behavior, what makes the curvature **κ** to a significant parameter for the shape description and the modelling of spectral changes.

Mathematically, the curvature is the change of a curve that occurs when the curve is traversed and can be expressed in parametric form for each point $P(x(t), y(t))$ using equation 1, where points refer to derivatives.

$$\varkappa = \frac{|\dot{x}\ddot{y} - \dot{y}\ddot{x}|}{(\dot{x}^2 + \dot{y}^2)^{\frac{3}{2}}} \quad (1)$$

While the curvature of a straight line is zero everywhere and the curvature of a circle is equal at all points, the curvature for all other curves changes from point to point and indicates how strongly the curve at a point $P$ deviates from a straight line. Thus, for the description of spectral-shapes, we use the following properties of curvature:

- The dimension of the current rate of change of the direction of a point moving on the curve. The greater the curvature, the greater the dimension of change.

---

[2] Plastic samples were kindly provided by PlasticsEurope Deutschland e.V. (https://www.plasticseurope.org/de).

- The behavior of the curvature. If the curvature value is positive, it is called a convex curve and in the case of a negative curvature value, it is considered a concave curve.

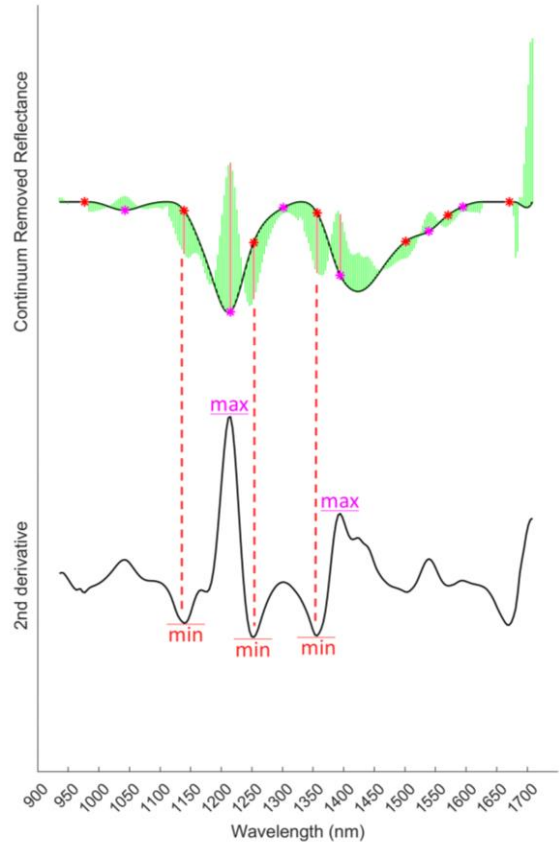

**Figure 2.** Use of minimum and maximum positions of 2nd derivative to select significant parameters for shape description. Parameters are the location (min/max), the curvature values at these locations (red lines) and the direction of curvature value (up for convex and down for concave behavior).

As shown in Figure **2**, extreme values of the second derivative are used to select significant parameters. The combination of these parameters with selected spectral values allows a precise description of the shape of spectral curves using a few selective spectral bands. The base for the calculation of the curvatures are preprocessed spectral curves. The preprocessing consists of a smoothing and a subsequent step of Continuum Removal.

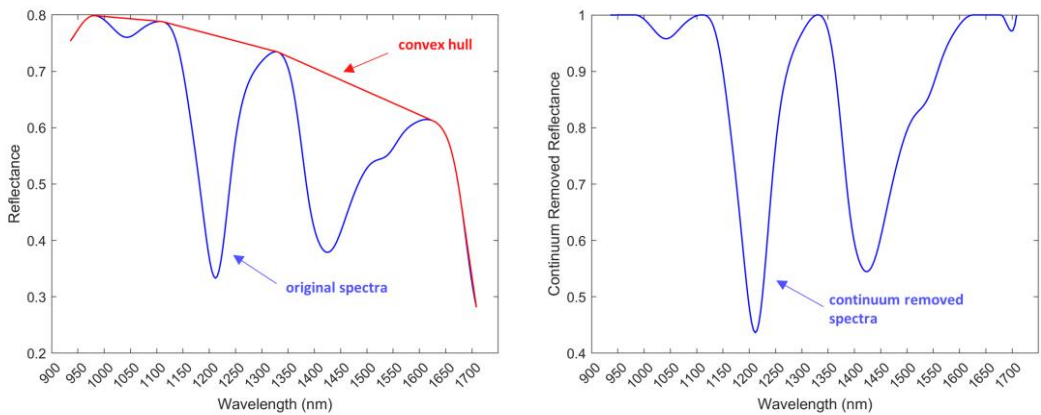

**Figure 3.** The use of Continuum Removal for quantification of absorption peaks by connecting the maxima of spectra with straight lines (left), setting the convex hull to 100% and subtracting the original spectra (right).

Continuum Removal is a normalization procedure which allows a better quantification of absorption peaks after removing the overall concave shape of spectral curves [38,39] and is illustrated in Figure **3**. Due to the particularly highlighted absorption bands, rules based on curvature values can be developed much more efficiently. An example for continuum removed spectra and the calculated curvatures is shown in Figure **4**. Depicted are continuum removed spectra for two different plastic types (PE, PS). In addition to the course of the spectral signature, the calculated curvature values are shown as positive and negative vertical lines in green and red color. The longer the line, the stronger the curvature at the respective band.

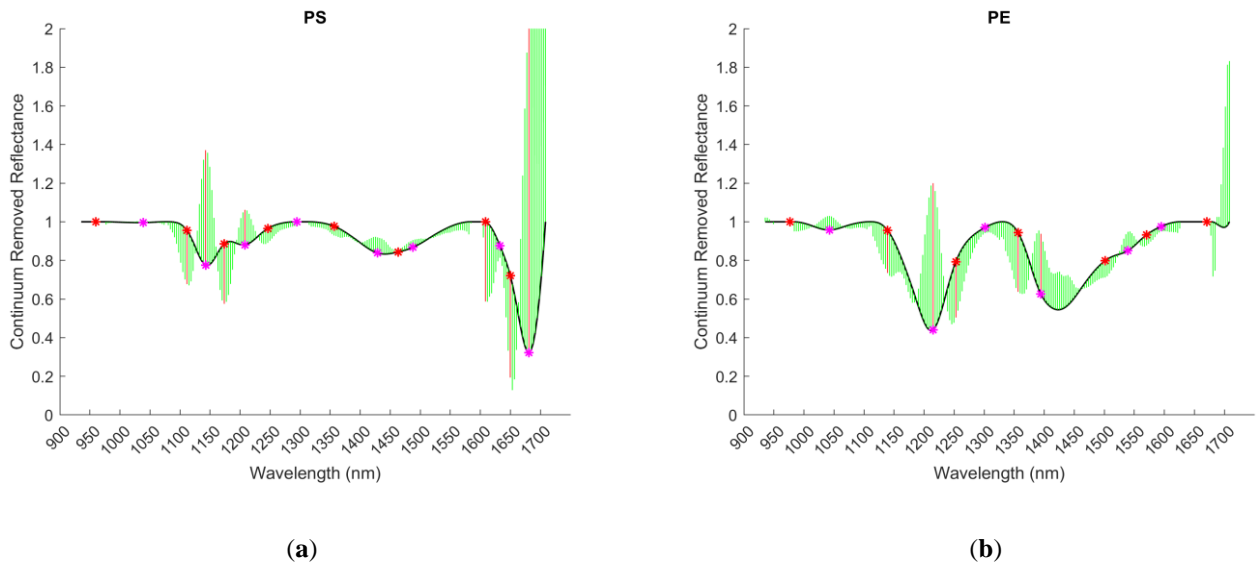

(**a**)            (**b**)

**Figure 4.** Continuum Removed Reflectance spectra (black line), calculated curvature values (green lines), maximum points with concave behavior (red dots) and minimum points with convex behavior (magenta dots) for (a) Polystyrene (PS); (b) Polyethylene (PE).

The curve behavior is represented by red and magenta colored dots. These points are maximum and minimum points and are automatically determined by the local maxima and minima of the second derivative. It helps to distinguish between concave and convex curve behavior, providing an important source of information for describing the shape. For the subsequent formation of rules, mainly the red lines are used, since these reflect both the concave and convex behavior and a significant change in curvature. All other curvature values (green lines) are not considered in the rule formation. To ensure that only significant changes in the curve are captured, a threshold is set for the selection of the relevant curvatures (red lines). The setting of the threshold mainly depends on the curve shape. The smaller the threshold the more detailed the description of the curves. However, for highly variant spectra (Figure **5**) higher threshold is sufficient. The examples in Figure **4** show the result of selected red lines for a threshold of 0.1. Building on these extracted parameters regarding the spectra for each material, a collection of conditions is formulated and used for the classification. It should be mentioned here that in addition to the used local shaped values, other rules can also be integrated. For example, rules describing global shape effects (e.g. the expressivity of the green peak for plants) or adding relations of averaged reflectivity in different regions as indices like NDVI do.

One of the main advantages of rule-based classifier is the simplicity. Once the knowledge on which rules are based has been worked out, conditions can be easily set up. Further advantages are the performance, the ability to handle redundant and irrelevant attributes and the flexible extensibility of rule sets [40]. Acquiring knowledge can seem effortful, but in view of the resulting advantages, it should be seen as a clear benefit which allows to structure and simplify problems by using expert knowledge. A comparison of deep learning and a knowledge-based method can be found in [41] and show, that a rule-based method can even be better as machine learning-based methods. For better illustration of rule formation, we refer to the spectral curves of PE and PS from Figure **4** and express it as shown in **Listing 1**.

**Listing 1:** Shape-based Rule for PS and PE.

| IF $CV_{1108}$ < −0.1 **AND** | IF $CV_{1139}$ < −0.1 **AND** |
| $CV_{1174}$ < −0.1 **AND** | $CV_{1253}$ < −0.1 **AND** |
| $CV_{1608}$ < −0.1 **AND** | $CV_{1357}$ < −0.1 **AND** |
| $CV_{1143}$ > +0.1 **AND** | $CV_{1215}$ > +0.1 **AND** |
| $CV_{1204}$ > +0.1 **AND** | $CV_{1394}$ > +0.1 **AND** |
| $CV_{1677}$ > +0.1 | **THEN** $class \rightarrow PE$ |
| **THEN** $class \rightarrow PS$ | |

The parameter *CV* stands for the curvature value at a specific spectral band and a positive or negative threshold is used to distinguish between convex and concave curve behavior. All conditions with a negative threshold will capture the downward red lines (concave behavior), while all conditions with a positive threshold will capture upward red lines (convex behavior). As already mentioned, for rule formation mainly the bands with high curvature are taken into account (red lines). Nevertheless, this does not mean that all bands are really necessary for the rule formation. By analyzing the data in advance, it is possible to achieve a clean classification even with a smaller number of selected bands. Therefore, only six conditions are defined for sample PS in **Listing 1**, while in Figure **4** a total of seven red lines are present. The continuum removed spectra and rules for all other existing plastic types in the used dataset are listed in Appendix A. The corresponding spectral signatures are also shown in Figure **5**. As an additional condition the continuum removed reflectance value (CRRV) could be used, if it is not possible to sufficiently distinguish the shape. For instance, to separate PF-black from all other available plastic types.

## 3. Results

Once the rules are established, the next step is to apply them. This involves a pixel by pixel processing of the corresponding dataset and a check of the rule conditions for each individual pixel. If a condition applies, this pixel will be assigned to the appropriate class.

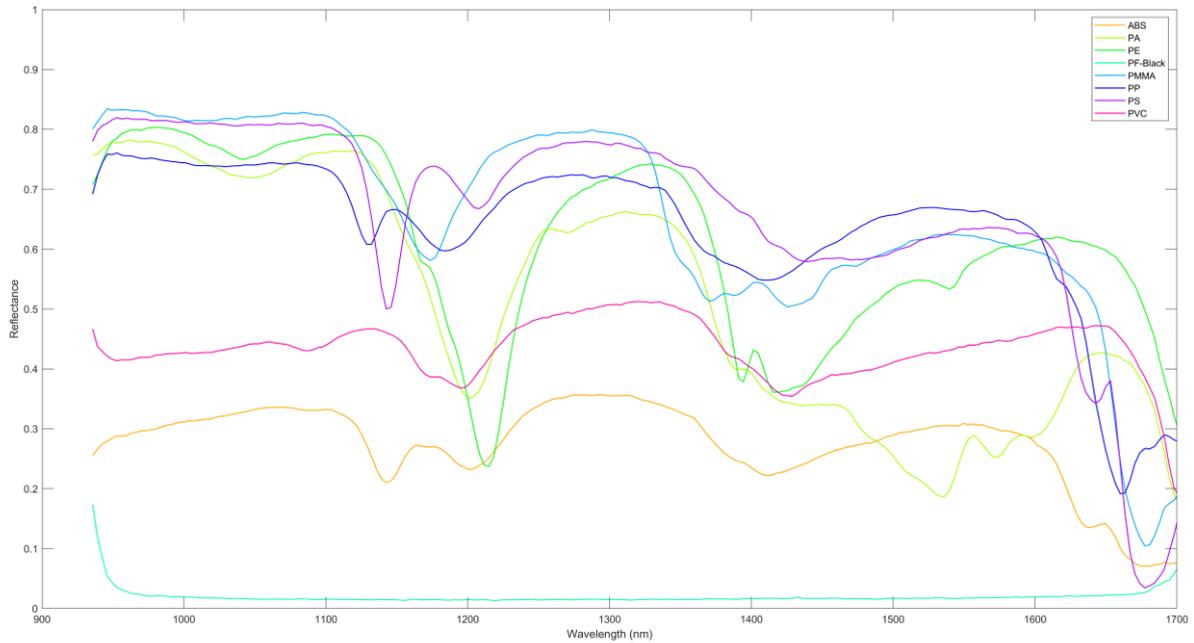

**Figure 5.** Reflectance spectra of available plastics. Each category of plastic material corresponds with a specific spectral shape, likewise the shapes of Continuum Removed Reflectance Spectra are different. This means, the rule defined for a category is specific to this category and there is no possible confusion as the shapes are distinct enough. So, such rules can be used only if spectral curve shapes are all sufficiently distinct.

## 3.1. Classification results for plastic samples

Applying the developed rules in Appendix A to the dataset consisting of plastic samples in Figure **1**, results shown in Figure **6** were obtained. The result of the classification shows that the individual samples were classified correctly. A closer look at the border areas of the samples shows that a shadow effect occurs and that this effect can influence the quality of the classification. This shadow effect occurs especially with samples (e.g. PS) that do not lie in a planar position.

Another point to consider here is the classification of PF samples. The continuum removed reflectance spectrum illustrated in Appendix A does not correspond to the real spectrum of the material PF. The PF samples used here are composed of black colorants. A well-known problem is the strong absorption of black colorants like carbon black. Due to the strong absorption of light from the UV to the NIR there is no reflected light that can be detected by the sensor and thus no spectral information that can be used for a classification [42,43]. The rule-based approach presented here offers the advantage that even in the absence of reflectance rules can be established based on the existing limited information, which permit such a classification. In the case of these samples, it means all black plastics will be classified as PF-Black. It must also be noted in this example that the samples used have clean, homogeneous surfaces and the resulting spectra have a correspondingly high degree of shape similarity. However, considering applications in the field of waste sorting, it is more common to work with dirty and damaged materials.

Therefore, for the evaluation of the methodology, a dataset (1253 x 578 pixels and 229 bands) with real waste consisting of plastic parts (objects 1 - 19) and circuit boards (objects 20 - 27) was processed additionally. The classification result for the real waste dataset is shown in Figure **7**. The objects in the dataset were chosen randomly from a collection of different plastic parts. Therefore, it is not surprising that certain materials (e.g. PMMA, PVC, PE and UP) are not present. In addition to the plastic types mentioned so far, polypropylene (PP) and acrylonitrile butadiene styrene (ABS) were identified on the basis of spectra from literature [44,45] and formulated as rules. Furthermore, an additional rule was established for the classification of printed circuit boards. A look at the classified plastic parts shows that the non-homogeneity of the surfaces is partly reflected in the results in form of unclassified or misclassified pixels. Nevertheless, it can be stated that the classification was generally successful by using the developed ten rules. The black plastic parts (objects 1 and 17 in Figure **7**) that are missing were not modelled in this example, because the used background also consists of black plastic and a separation in this special case proves to be difficult. Also, since the spectra of object number 5 in Figure **7** could not be assigned to any material, no rule for classification was established.

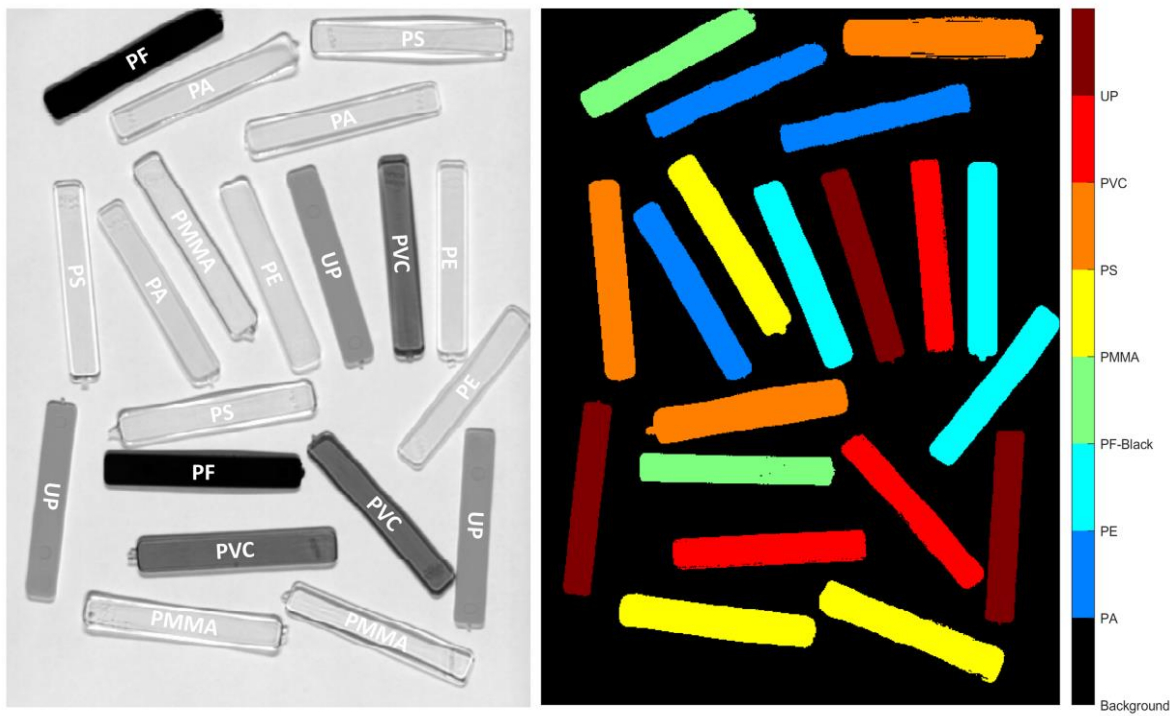

**Figure 6.** Classification result using developed rules.

An important step in the recycling process is the separation of plastics and electronic waste. In particular, the high variation of different material compositions from which PCBs are made makes separation difficult. However, the example shown here also demonstrates that prior analysis of data and the use of rules based on acquired knowledge can lead to a more efficient recycling process. Due to the already mentioned high variation of PCBs and the composition of different materials (e.g. board, conductors, solder joints, resistors, capacitors), a holistic assessment of a PCB is difficult to implement.

Nevertheless, it is evident in this example that the board has been identified in all cases and, as expected, only the areas with conductive tracks and metallic or electronic components have not been assigned and consequently ended up in a category with black plastics and the background. A significant difference to the samples shown in Figure **6** is the heterogeneity of the surfaces. Differences in depth, shadow effects on the surface and dirt lead to a very high variability of spectral signatures. As an example, a region of object number 12 in Figure **7** was selected. In this area, differences in depth, shadow effects and soiling can be found. A representation of the high variability is shown in Figure **8**. The shape-based classification result for object number 13, however, shows that a satisfactory result was achieved despite the high variability.

The processing time using MATLAB on a machine with an Intel(R) Core(TM) i7-10700 CPU @ 2.90GHz and 16 GB RAM was 136 seconds for the real waste dataset and 44 seconds for the plastic dataset from Figure **6**.

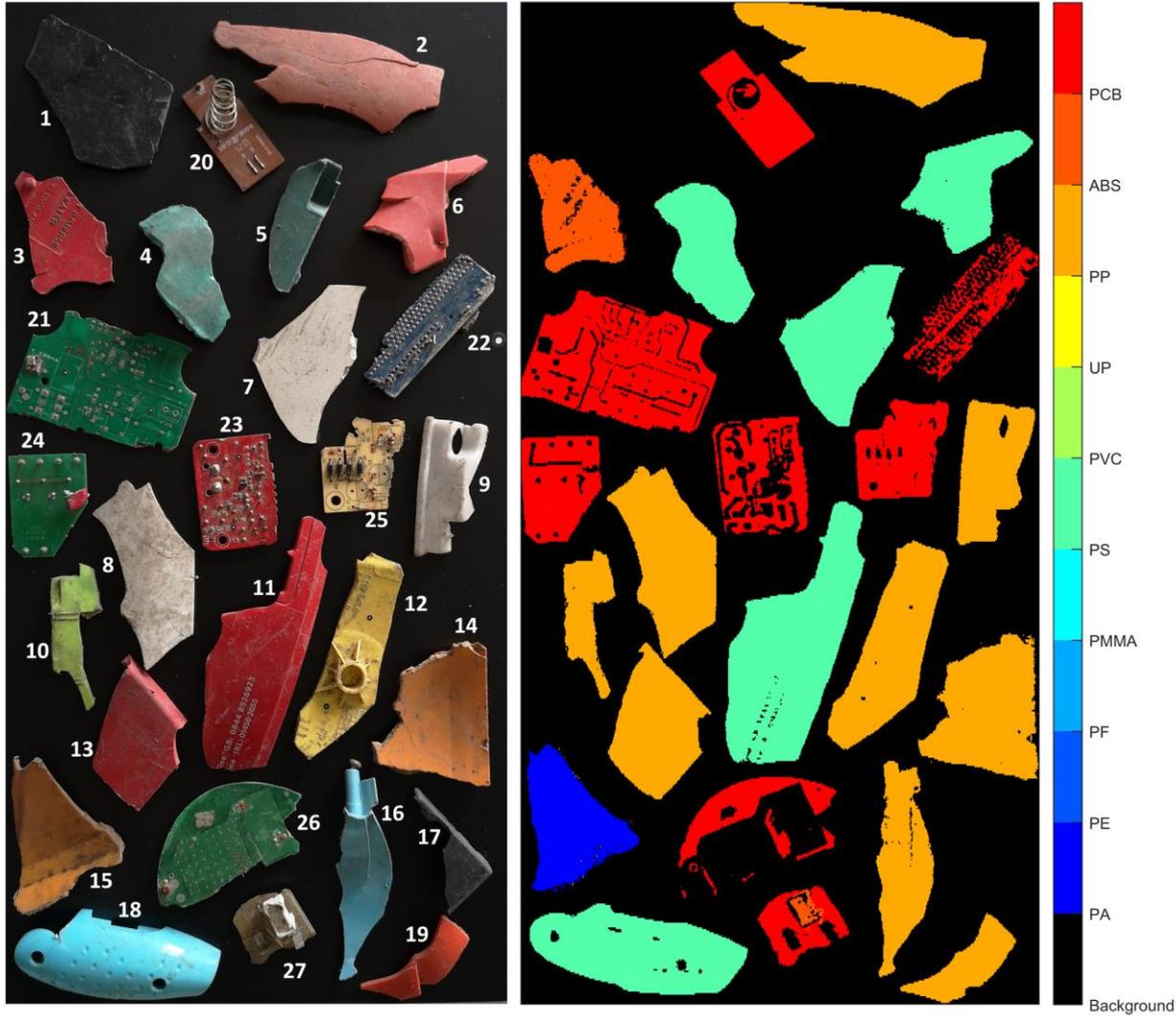

**Figure 7.** Classification result for real waste (plastics and printed circuit boards) using developed rules.

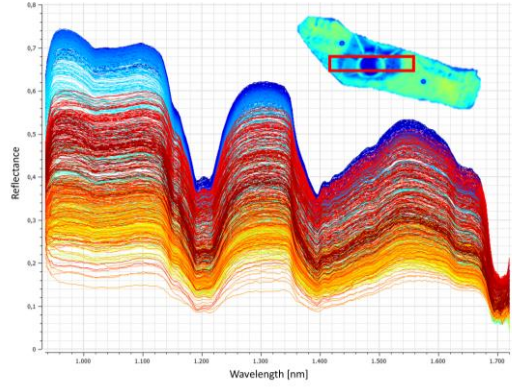

**Figure 8.** Spectral signature variability for object 12 of real waste dataset (spectra of all pixels inside the red box).

In order to compare the quality of the proposed method to standard methods, the dataset shown in Figure **6** (plastic samples) and the dataset shown in Figure **7** (real waste samples) were also processed using a supervised multiclass SVM classifier (C-SVC, One-Versus-All), which is based on the selection of a training set to obtain a model and the subsequent application of this model for the prediction of classes. The dataset with the real waste was limited to the plastic objects. As already described in [17], the quality of the results strongly depends on the chosen training samples and parameterization. For this reason, three different kernels, namely Radial Basis Function (RBF), linear and polynomial, with different parameters have been tested on the datasets. Due to the multicategory classification, a classification threshold is used, which is the maximum distance to the hypersurface for conducted classification to a specific class. The parameters for the kernels and the classification threshold were obtained by testing and checking the results. Because of the homogeneity of the plastic samples, one material sample of each material class was manually labelled and used as training set. A description of the datasets can be found in Table **1**. The best results for the plastic dataset using different kernels and classification thresholds are shown in Figure **9**.

The accuracy of each classification method compared is reported in Table **2**. In principle, the results of this homogeneous and optimal dataset are comparable for most of classification metrics used. Nevertheless, significant differences can be seen in the resulting images, especially at the edge of objects, in form of unclassified or misclassified pixels. As already mentioned, this is due to a shadow effect that locally occurs in some areas, which impacts mainly the quality of the SVM methods. While the rule-based approach also classifies these edge areas and some shadow parts as corresponding material, the SVM results in misclassifications, especially using linear and polynomial kernels.

**Table 1.** Number of samples and training samples for each class.

|  | Class | Number of Samples | Number of Training Samples |
|---|---|---|---|
| **Plastic dataset** | PA | 16276 | 3783 |
|  | PE | 16212 | 3659 |
|  | PF-Black | 11157 | 4930 |
|  | PMMA | 15779 | 4071 |
|  | PS | 16877 | 4123 |
|  | PVC | 16645 | 4869 |
|  | UP | 16098 | 4613 |
| **Real waste dataset** | PA | 13343 | 2114 |
|  | PS | 81238 | 10444 |
|  | PP | 116620 | 11936 |
|  | ABS | 11703 | 1638 |

**Table 2.** Classification accuracies for plastic dataset.

| Metrics | Rule-based Method | SVM (a) | SVM (b) | SVM (c) |
|---|---|---|---|---|
| Overall Accuracy | 0.9694 | 0.9647 | 0.9641 | 0.9641 |
| Precision | 0.9604 | 0.9731 | 0.9403 | 0.9403 |
| Sensitivity | 0.9551 | 0.9272 | 0.9623 | 0.9623 |
| False Positive Rate | 0.0077 | 0.0114 | 0.0076 | 0.0076 |
| F1-Score | 0.9564 | 0.9487 | 0.9503 | 0.9503 |
| Kappa | 0.8603 | 0.8384 | 0.8358 | 0.8359 |

(a) Solver: C-SVC; Classification Type: One-Versus-All; Kernel: RBF; C-Value: 1; Threshold: 0.5.
(b) Solver: C-SVC; Classification Type: One-Versus-All; Kernel: linear; C-Value: 1; Threshold: 0.3.
(c) Solver: C-SVC; Classification Type: One-Versus-All; Kernel: polynomial; C-Value: 1; Threshold: 0.3.

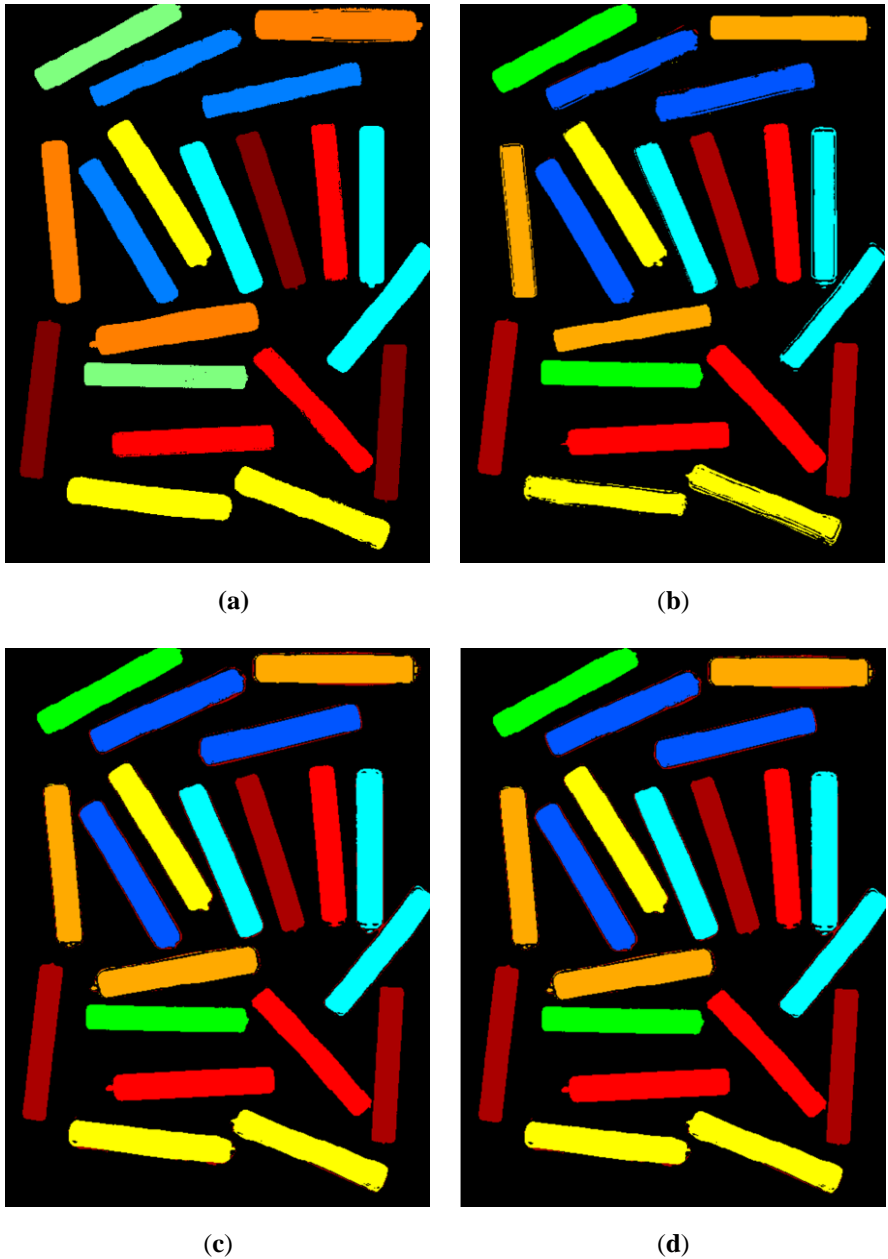

**Figure 9.** Classification results using (a) proposed shape-based method; SVM with (b) RBF kernel, C-Value of 1 and classification threshold of 0.5; (c) linear kernel, C-Value of 1 and classification threshold of 0.3; (d) polynomial kernel, C-Value of 1 and classification threshold of 0.3.

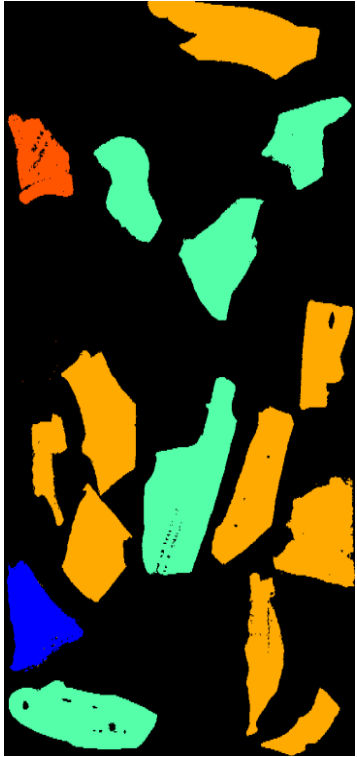 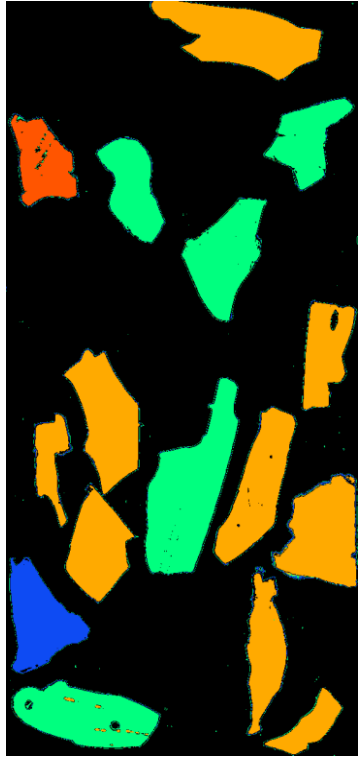 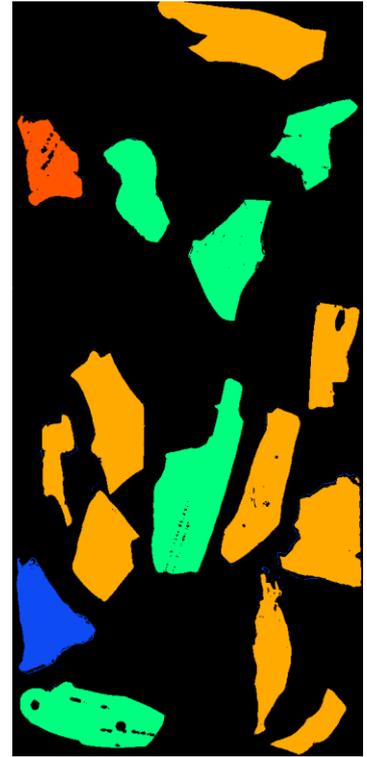
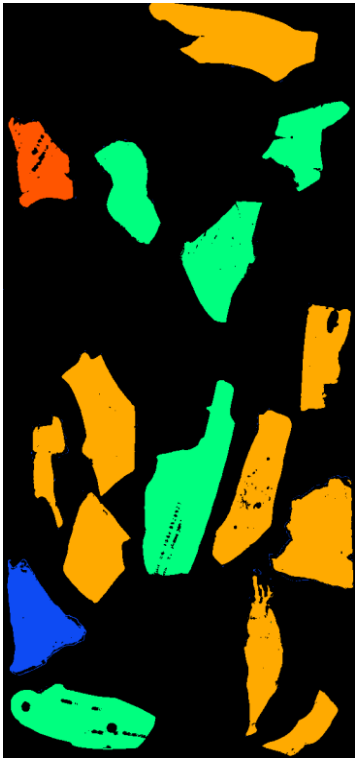 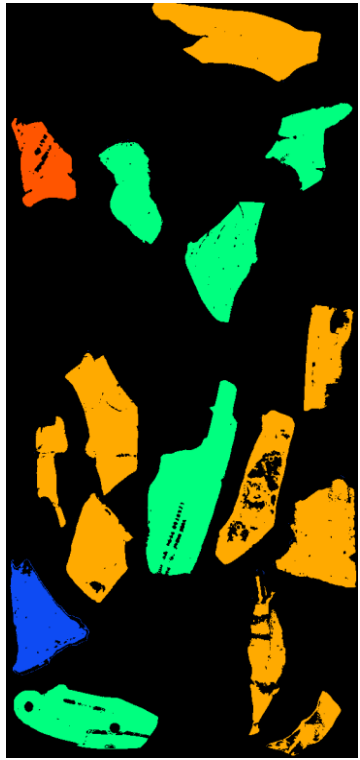 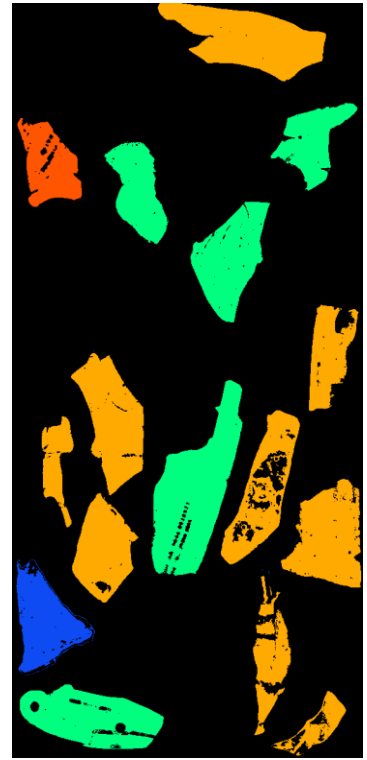

**Figure 10.** Classification results using (a) proposed shape-based method; SVM with (b) RBF kernel, C-Value of 0.2 and classification threshold of 0.2; (c) RBF kernel, C-Value of 0.2 and classification threshold of 0.3; (d) RBF kernel, C-Value of 0.2 and classification threshold of 0.4; (e) RBF kernel, C-Value of 0.2 and classification threshold of 0.5; (f) RBF kernel, C-Value of 0.2 and classification threshold of 0.6.

**Table 3.** Classification results for real waste dataset.

| Metrics | Rule-based Method | SVM (a) | SVM (b) | SVM (c) | SVM (d) | SVM (e) |
|---|---|---|---|---|---|---|
| Overall Accuracy | 0.9842 | 0.9770 | 0.9826 | 0.9805 | 0.9760 | 0.9586 |
| Precision | 0.9842 | 0.9398 | 0.9698 | 0.9807 | 0.9856 | 0.9854 |
| Sensitivity | 0.9498 | 0.9477 | 0.9434 | 0.9353 | 0.9208 | 0.8831 |
| False Positive Rate | 0.0075 | 0.0093 | 0.0098 | 0.0119 | 0.0152 | 0.0267 |
| F1-Score | 0.9661 | 0.9396 | 0.9540 | 0.9556 | 0.9504 | 0.9290 |
| Kappa | 0.9506 | 0.9283 | 0.9456 | 0.9392 | 0.9249 | 0.8707 |

(d) Solver: C-SVC; Classification Type: One-Versus-All; Kernel: RBF; C-Value: 0.2; Threshold: 0.2.
(e) Solver: C-SVC; Classification Type: One-Versus-All; Kernel: RBF; C-Value: 0.2; Threshold: 0.3.
(f) Solver: C-SVC; Classification Type: One-Versus-All; Kernel: RBF; C-Value: 0.2; Threshold: 0.4.
(g) Solver: C-SVC; Classification Type: One-Versus-All; Kernel: RBF; C-Value: 0.2; Threshold: 0.5.
(h) Solver: C-SVC; Classification Type: One-Versus-All; Kernel: RBF; C-Value: 0.2; Threshold: 0.6.

Compared to the plastic dataset, the real waste dataset has a typical situation consisting of objects with a high degree of variety. This not only increases the effort required to train the model, but also makes the training process more difficult. Small subsets of the individual samples were manually selected as training set. This strategy was considered to be appropriate due to the strong differences between the plastic samples. Thus, when selecting the subsets, special care was taken to ensure that particularly critical areas, such as shadow areas caused by depth, dirty areas and damaged surfaces, were also covered in the training set. The best resulting class-labels are presented in Figure **10** and show also the effect of different classification thresholds. The corresponding numerical values are listed in Table **3**.

In principle, also for this dataset it can be stated that, despite the challenging objects of the real waste dataset, a high degree of accuracy can be achieved for both methods. The best SVM result was obtained with an RBF kernel consisting of a C-Value of 0.2 and a classification threshold of 0.3. In comparison, the rule-based method produces slightly better values for the different metrics. Also noticeable is the more homogeneous representation of objects in the resulting class-labels in Figure **10**. One of the advantages of the shape-based approach is that, theoretically, only one spectrum per material is required for the rule formation. When using an SVM, it is necessary to ensure that in case of high variability datasets, all data representing the spectral variations of one material class are included in the training set. Fulfilling this requirement can be a challenging task.

### 3.2. Classification results for plants

Compared to the dataset consisting of plastics and electronic waste, spectral signatures of plants often show a very similar curve shape. This fact generally complicates the classification process.

To illustrate the ability of classifying even in such difficult cases and the flexibility of the methodology presented in this work, rules were developed to distinguish between plant species based on the spectral signatures shown in Figure **11**, which are coherent with spectral signatures provided in [46]. Due to the similarity of the shape, in this particular case, rules have been formulated using mainly Continuum Removed Reflectance Values (CRRV) as well as curvature parameters. This involved using the entire spectrum from 500 nm to 1700 nm to work out fine differences regarding the water content, the cell structure and the pigments of used plants. In this context, it must be taken into account that the spectral properties of a plant species can also vary, as spectral signatures will be affected by factors such as species, variety, age, internal cell structure, environmental conditions, chemical composition and nutrient content [47]. An example for such a variation is given in Figure **12** for moss. Although not implemented in this work, one possible way to deal with such heterogeneous spectra could be the addition of rules based on texture features or other local image features [48–50].

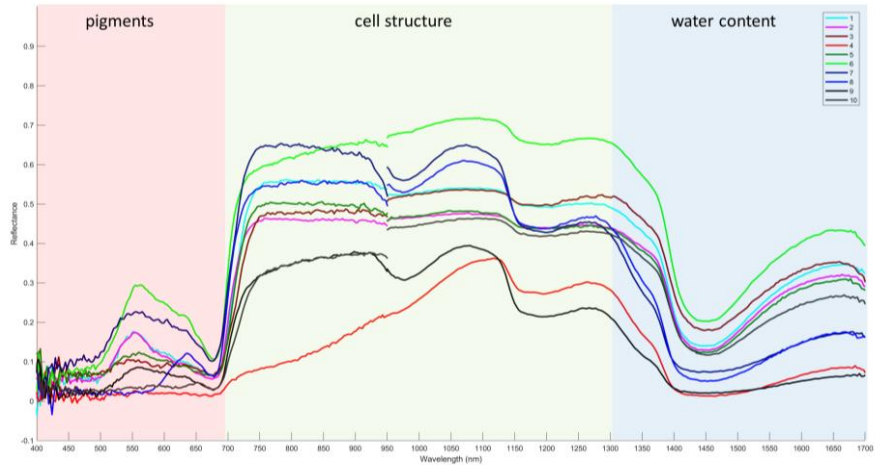

**Figure 11.** Spectra of measured plants. Two different cameras were used to acquire the spectra (Specim FX10 and Specim FX17). The discontinuities in the range of 900 nm and the greater noise in the VIS range can be explained by the differences in spectral sensitivity of the cameras. In particular, the range below 500 nm exhibits strong noise effects and was not considered in the rule formation. Taraxacum (1), inula conyza (2), campanula rotundifolia (3), moss (4), geranium robertianum (5), asplenium trichomanes (6), green sedum (7), red sedum (8), moss (9), lichen (10).

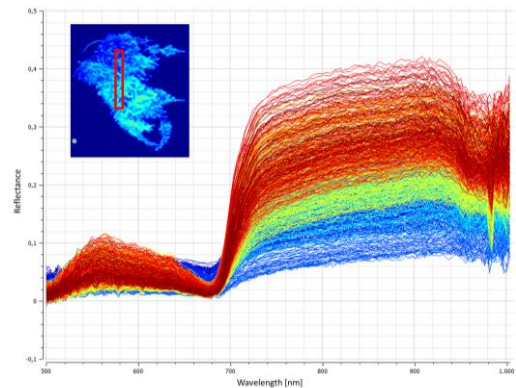

**Figure 12.** Spectral variations in VIS for one type of moss regarding all pixels in the red box.

A total of six rules were developed to separate the ten samples in groups of moss, lichen, red sedum, green sedum, geranium robertianum and green leaves. The results in Figure **13** show that, in principle, a classification between different plant species is possible. For example, it can be clearly seen that red sedum is distinctly different from the other plant species due to its coloring. Green sedum also differs based on the cell structure, whereby the properties of this plant seems also partially reflected in moss, which can be relatively well distinguished from the other plants due to the spectral signatures in the NIR and the low reflectance around 550 nm. Furthermore, it was possible to classify the lichens in object 10.

The very strong spectral similarity between the leaves does not allow a clean separation between the individual species. Only geranium robertianum differs due to its low reflectance in the range around 550 nm. In this context, it is important to note that the spectral signatures shown in Figure **11** only reflect the spectral signature of one pixel and that the recognizable differences, refer to reflectance values or shape, are not consistently present due to the high variability within a plant species. Nevertheless, it can be stated also for this example that a rule-based approach is a flexible method and - depending on prior analytical efforts - has the potential to provide useful results.

**4. Discussion**

Results of a rule-based classification approach were presented, based on the shape of spectral signatures. The rules are established in a supervised way and are not only based on spectral values, but especially on parameters that describe the geometric form of a spectral signature. These parameters include the automatic determined curvature points (i.e. specific spectral values obtained through the $2^{nd}$ derivative), curvature values and curvature behavior. The effectiveness of this method was demonstrated with different datasets from completely different fields of application. In particular, the separation of materials with significant geometric differences in the course of the curve lead to convincing results.

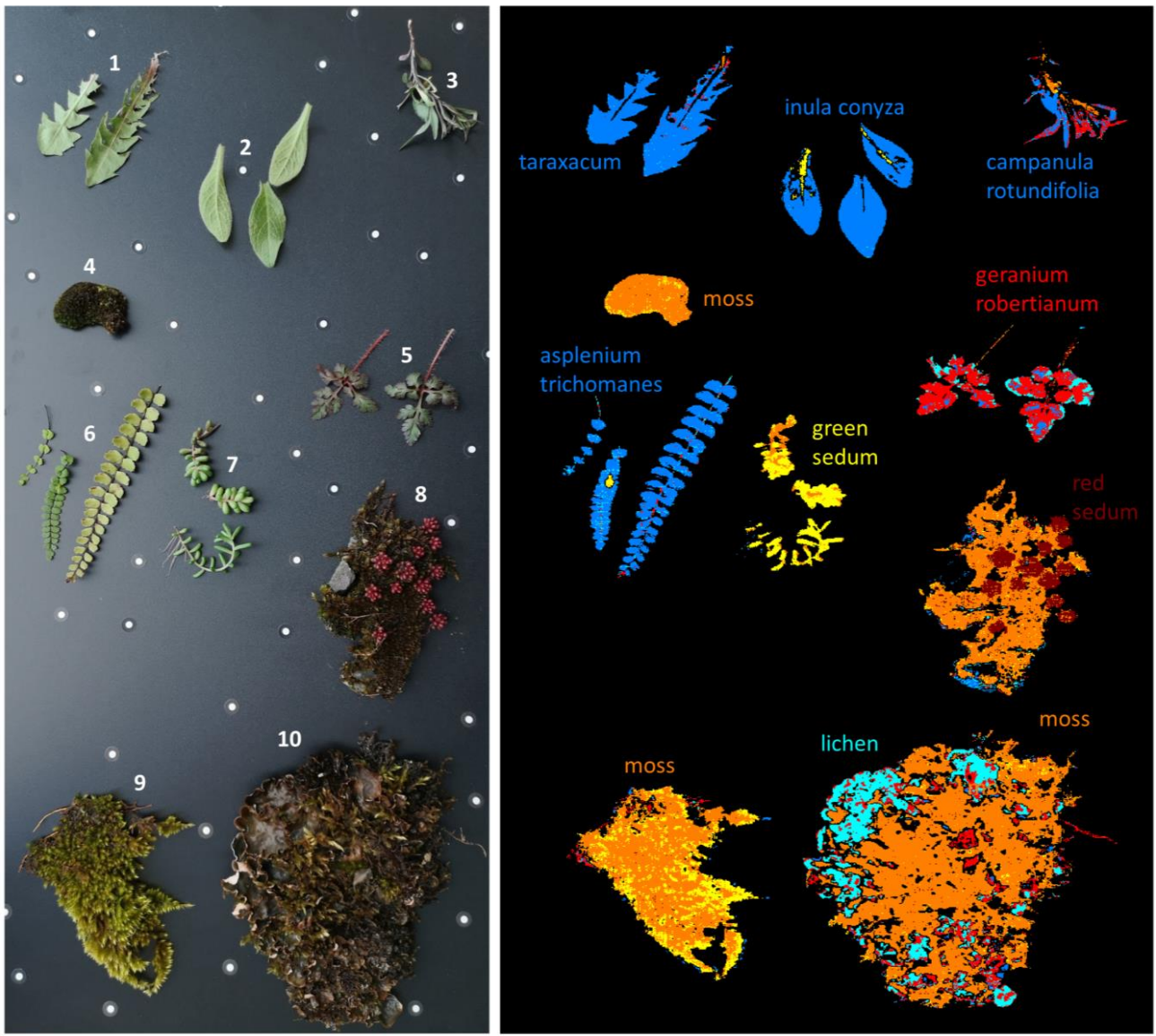

**Figure 13.** Classification result for plants using VIS and NIR (500 nm – 1700 nm).

An essential advantage over classical classification approaches is the possibility of describing the course of a spectrum in detail on the basis of a few selective parameters. Classic approaches do not require a previous analysis of the data and offer the advantage of automated processing based on the actual data. A precondition for this, however, is a dimensional reduction. This reduction is a selection of informative bands and often leads to a loss of information within the spectral signatures, because it is based on pure statistical reduction. With the method presented here, the finest changes in the course of the curve can be identified and modelled. The prerequisite for this is a previous analysis of the spectral signatures and the establishment of the rules based on expert knowledge. This process may be considered time consuming, but considering the parameters used, it does not require a particularly high investment of time. An automated generation of the rules would also be conceivable, since the essentially used parameters such as curvature value, curvature point and curvature behavior are determined by simple mathematical methods. A disadvantage that can arise with the automatic formulation of rules is that the number of rules increases with the number of categories. This can lead to complex rules that conflict with each other. Modelling the rules based on expert knowledge can avoid this by understanding the interaction of the factors, which makes it possible to formulate the simplest concept that gives the best results. Basically, when dealing with spectral data, it can be stated that the number of categories that can be classified from spectral bands is small enough that a limited number of rules can be used.

The analysis of spectral signatures in advance has the further advantage that there is no need to use the entire spectrum for classification. If significant shape differences are detected within a limited spectral range, it is sufficient to consider only this range in order to establish the rules based on it.

While spectral values can vary strongly depending on different factors, differences in the shape of spectral signatures are only due to differences in material composition. This fact is the reason for the promising results, which confirm our expectations of this basic idea of using the individual fingerprint of different materials or objects.

Furthermore, the processing time must be mentioned. The pixel-by-pixel processing and checking for conditions takes 113.42 seconds for the plant dataset (1220 x 640 pixels and 462 bands) shown in Figure **13.** Classification result for plants using VIS and NIR (500 nm – 1700 nm). In comparison, the shape-based classification in [30] for a dataset of 512 x 512 pixels and 6 bands takes about 9 minutes (computer configuration: CPU 2.93 GHz and installed memory 4.00 GB). Another comparison with a research using a classic classification method like SVM also shows a better performance. The processing time for a comparable dataset of 1168 x 696 pixels and 520 bands takes 2980.21 seconds using 30 bands and 9381.37 seconds using all bands of the dataset (computer configuration: Intel Core i7-6800k 3.40-GHZ CPU and installed memory 64 GB) [51].

## 5. Conclusion

Considering the shape of spectral signatures and describing them by curvature parameters and spectral values proves to be a convincing supervised method for classifying diverse groups of materials from different fields of application. Further evaluation on publicly available datasets, e.g. from the field of remote sensing, also with respect to other classification methods, would be of interest. A possible disadvantage of this methodology could be the necessary prior analysis of the data. This circumstance could be eliminated by automation. Especially in the case of significantly different spectra, like the plastic samples, automation is certainly feasible and would bring the benefit of an unsupervised method.

**Funding:** This research was funded by the European Union from the European Regional Devel-opment Fund and the state of Rhineland-Palatinate.

**Acknowledgments:** We thank Pellenc ST for supplying the real waste materials.

# Appendix A

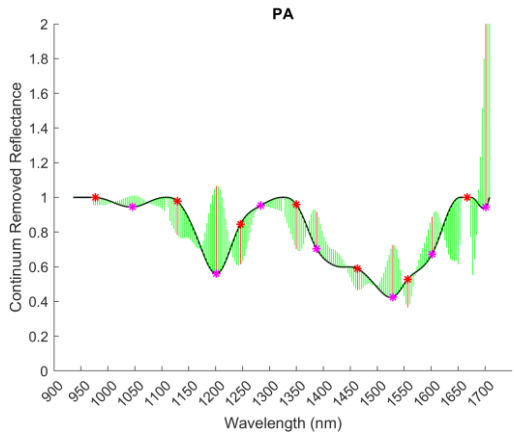

IF $CV_{1135} < -0.1$ **AND**
$CV_{1246} < -0.1$ **AND**
$CV_{1350} < -0.1$ **AND**
$CV_{1460} < -0.1$ **AND**
$CV_{1201} > +0.1$ **AND**
$CV_{1388} > +0.1$ **AND**
$CV_{1529} > +0.1$ **AND**
$CV_{1604} > +0.1$ **AND**
**THEN** $class \rightarrow PA$

**Figure A1**. Continuum removed reflectance spectra and shape-based rule for PA.

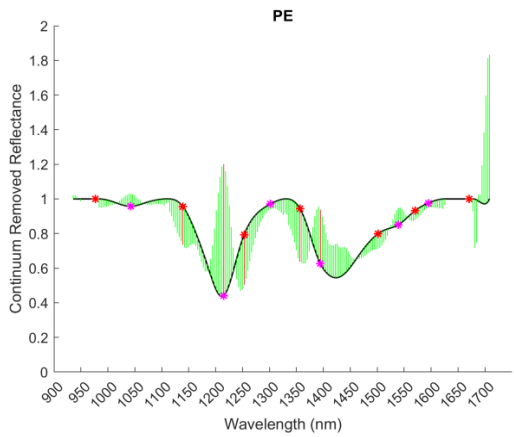

IF $CV_{1139} < -0.1$ **AND**
$CV_{1253} < -0.1$ **AND**
$CV_{1357} < -0.1$ **AND**
$CV_{1215} > +0.1$ **AND**
$CV_{1394} > +0.1$ **AND**
**THEN** $class \rightarrow PE$

**Figure A2.** Continuum removed reflectance spectra and shape-based rule for PE.

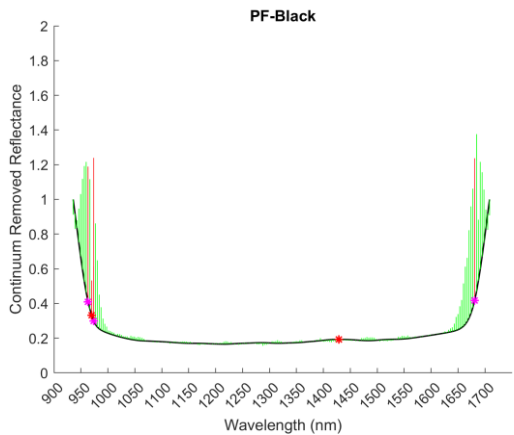

IF $CV_{973} > +0.1$ **AND**
$CV_{1681} > +0.1$ **AND**
$CRRV_{1429} < 0.6$
**THEN** $class \rightarrow \text{PF} - \text{black}$

**Figure A3.** Continuum removed reflectance spectra and shape-based rule for PF-black.

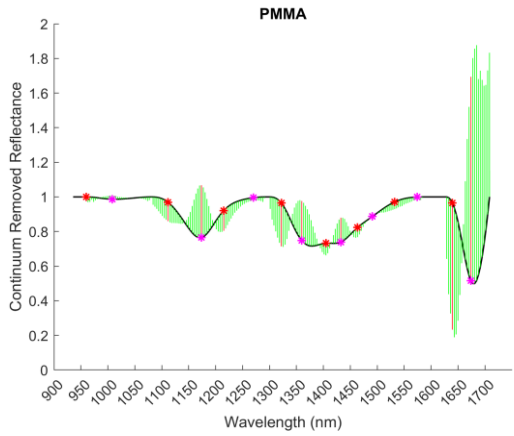

**Figure A4.** Continuum removed reflectance spectra and shape-based rule for PMMA.

IF $CV_{1322} < -0.1$ **AND**
$CV_{1639} < -0.1$ **AND**
$CV_{1174} > +0.1$ **AND**
$CV_{1360} > +0.1$ **AND**
$CV_{1432} > +0.1$ **AND**
$CV_{1674} > +0.1$ **AND**
**THEN** $class \to PMMA$

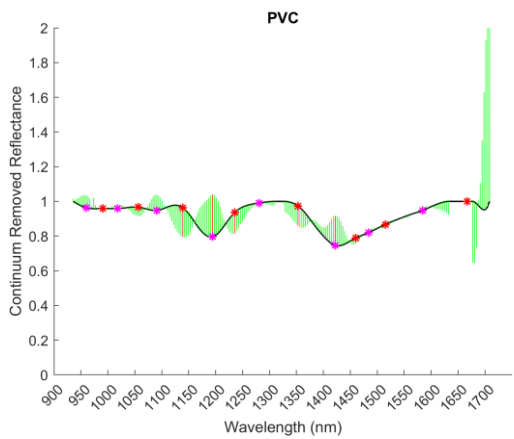

**Figure A5.** Continuum removed reflectance spectra and shape-based rule for PVC.

IF $CV_{1139} < -0.1$ **AND**
$CV_{1236} < -0.1$ **AND**
$CV_{1353} < -0.1$ **AND**
$CV_{1194} > +0.1$ **AND**
$CV_{1422} > +0.1$ **AND**
**THEN** $class \to PVC$

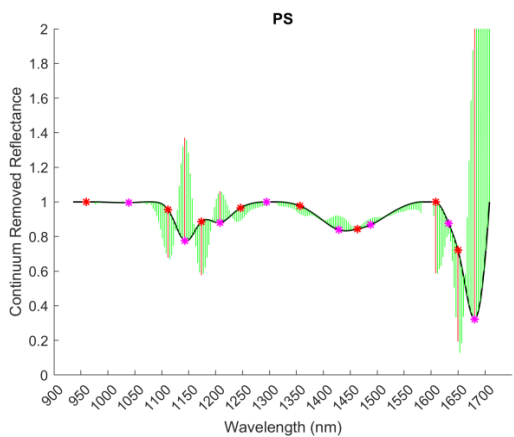

**Figure A6.** Continuum removed reflectance spectra and shape-based rule for PS.

IF $CV_{1108} < -0.1$ **AND**
$CV_{1174} < -0.1$ **AND**
$CV_{1608} < -0.1$ **AND**
$CV_{1143} > +0.1$ **AND**
$CV_{1204} > +0.1$ **AND**
$CV_{1677} > +0.1$
**THEN** $class \to PS$

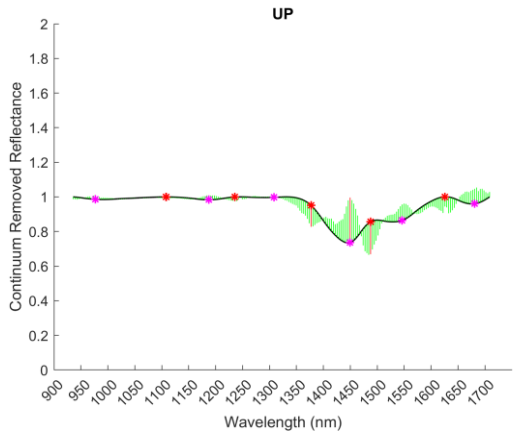

**Figure A7.** Continuum removed reflectance spectra and shape-based rule for UP.

IF $CV_{1377}$ < −0.1 **AND**
$CV_{1488}$ < −0.1 **AND**
$CV_{1450}$ > +0.1 **AND**
**THEN** $class \rightarrow$ UP

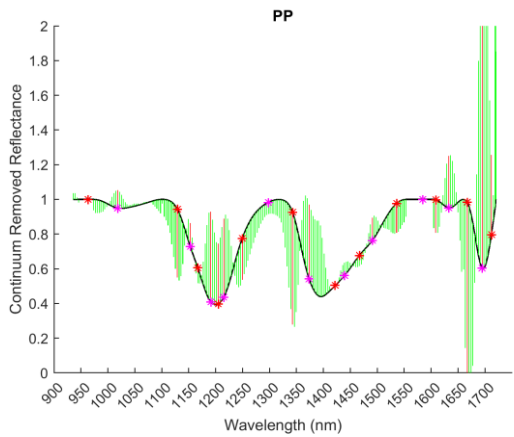

**Figure A8.** Continuum removed reflectance spectra and shape-based rule for PP.

IF $CV_{1128}$ < −0.1 **AND**
$CV_{1342}$ < −0.1 **AND**
$CV_{1190}$ > +0.1 **AND**
$CV_{1215}$ > +0.1 **AND**
$CV_{1387}$ > +0.1 **AND**
$CV_{1694}$ > +0.1 **AND**
**THEN** $class \rightarrow$ PP

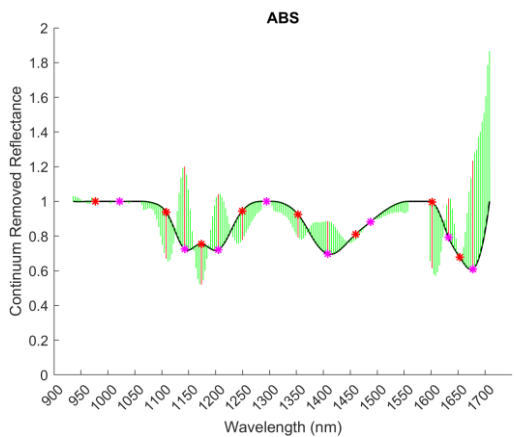

**Figure A9.** Continuum removed reflectance spectra and shape-based rule for ABS.

IF $CV_{1104}$ < −0.1 **AND**
$CV_{1152}$ < −0.1 **AND**
$CV_{1339}$ > −0.1 **AND**
$CV_{1629}$ > −0.1 **AND**
$CV_{1128}$ > +0.1 **AND**
$CV_{1187}$ > +0.1 **AND**
$CV_{1415}$ > +0.1 **AND**
$CV_{1656}$ > +0.1 **AND**
**THEN** $class \rightarrow$ ABS

## References

[1] M. Khan, H. Khan, A. Yousaf, K. Khurshid, A. Abbas, Modern Trends in Hyperspectral Image Analysis: A Review, IEEE Access 6 (2018) 14118–14129. https://doi.org/10.1109/ACCESS.2018.2812999.

[2] E. Honkavaara, M.A. Eskelinen, I. Polonen, H. Saari, H. Ojanen, R. Mannila, C. Holmlund, T. Hakala, P. Litkey, T. Rosnell, N. Viljanen, M. Pulkkanen, Remote Sensing of 3-D Geometry and Surface Moisture of a Peat Production Area Using Hyperspectral Frame Cameras in Visible to Short-Wave Infrared Spectral Ranges Onboard a Small Unmanned Airborne Vehicle (UAV), IEEE Trans. Geosci. Remote Sensing 54 (2016) 5440–5454. https://doi.org/10.1109/TGRS.2016.2565471.

[3] P. Vermeulen, P. Flémal, O. Pigeon, P. Dardenne, J. Fernández Pierna, V. Baeten, Assessment of pesticide coating on cereal seeds by near infrared hyperspectral imaging, Journal of Spectral Imaging 6 (2017). https://doi.org/10.1255/jsi.2017.a1.

[4] R. Roscher, J. Behmann, A.-K. Mahlein, J. Dupuis, H. Kuhlmann, L. Plümer, Detection of disease symptoms on hyperspectral 3D plant models, ISPRS Annals of Photogrammetry, Remote Sensing and Spatial Information Sciences III-7 (2016) 89–96. https://doi.org/10.5194/isprs-annals-III-7-89-2016.

[5] M. Al-Sarayreh, M. M. Reis, W. Qi Yan, R. Klette, Detection of Red-Meat Adulteration by Deep Spectral–Spatial Features in Hyperspectral Images, Journal of Imaging 4 (2018) 63. https://doi.org/10.3390/jimaging4050063.

[6] I. Pineda, N. Md, O. Gwun, Calyx and Stem Discrimination for Apple Quality Control Using Hyperspectral Imaging, 2019, pp. 274–287.

[7] Q. Wang, K.-Q. Yu, Rapid and Nondestructive Classification of Cantonese Sausage Degree Using Hyperspectral Images, Applied Sciences 9 (2019) 822. https://doi.org/10.3390/app9050822.

[8] M. Halicek, H. Fabelo, S. Ortega, G.M. Callico, B. Fei, In-Vivo and Ex-Vivo Tissue Analysis through Hyperspectral Imaging Techniques: Revealing the Invisible Features of Cancer, Cancers (Basel) 11 (2019). https://doi.org/10.3390/cancers11060756.

[9] H. Ding, R. C. Chang, Hyperspectral Imaging With Burn Contour Extraction for Burn Wound Depth Assessment, Journal of Engineering and Science in Medical Diagnostics and Therapy 1 (2018). https://doi.org/10.1115/1.4040470.

[10] M. Wahabzada, M. Besser, M. Khosravani, M. Kuska, K. Kersting, A.-K. Mahlein, E. Stuermer, Monitoring wound healing in a 3D wound model by hyperspectral imaging and efficient clustering, PLOS ONE 12 (2017) e0186425. https://doi.org/10.1371/journal.pone.0186425.

[11] A. Polak, T. Kelman, P. Murray, S. Marshall, D.J. Stothard, N. Eastaugh, F. Eastaugh, Hyperspectral imaging combined with data classification techniques as an aid for artwork authentication, Journal of Cultural Heritage 26 (2017) 1–11. https://doi.org/10.1016/j.culher.2017.01.013.

[12] F. Daniel, A. Mounier, J. Pérez-Arantegui, C. Pardos, N. Prieto-Taboada, S. Fdez-Ortiz de Vallejuelo, K. Castro, Hyperspectral imaging applied to the analysis of Goya paintings in the Museum of Zaragoza (Spain), Microchemical Journal 126 (2016) 113–120. https://doi.org/10.1016/j.microc.2015.11.044.

[13] C. Cucci, J.K. Delaney, M. Picollo, Reflectance Hyperspectral Imaging for Investigation of Works of Art: Old Master Paintings and Illuminated Manuscripts, Acc. Chem. Res. 49 (2016) 2070–2079. https://doi.org/10.1021/acs.accounts.6b00048.

[14] A. Majda, R. Wietecha-Posłuszny, A. Mendys, A. Wójtowicz, B. Łydżba-Kopczyńska, Hyperspectral imaging and multivariate analysis in the dried blood spots investigations, Applied Physics A 124 (2018). https://doi.org/10.1007/s00339-018-1739-6.

[15] S. Cadd, B. Li, P. Beveridge, W. T O'hare, M. Islam, Age Determination of Blood-Stained Fingerprints Using Visible Wavelength Reflectance Hyperspectral Imaging, Journal of Imaging 4 (2018). https://doi.org/10.3390/jimaging4120141.

[16] G. Edelman, E. Gaston, T. van Leeuwen, P. Cullen, M. Aalders, Hyperspectral Imaging for Non-Contact Analysis of Forensic Traces, Forensic science international 223 (2012). https://doi.org/10.1016/j.forsciint.2012.09.012.

[17] P. Ghamisi, J. Plaza, Y. Chen, J. Li, A.J. Plaza, Advanced Spectral Classifiers for Hyperspectral Images: A review, IEEE Geosci. Remote Sens. Mag. 5 (2017) 8–32. https://doi.org/10.1109/MGRS.2016.2616418.

[18] J. Ren, R. Wang, G. Liu, R. Feng, Y. Wang, W. Wu, Partitioned Relief-F Method for Dimensionality Reduction of Hyperspectral Images, Remote Sensing 12 (2020) 1104. https://doi.org/10.3390/rs12071104.

[19] S. Li, Q. Hao, X. Kang, J.A. Benediktsson, Gaussian Pyramid Based Multiscale Feature Fusion for Hyperspectral Image Classification, IEEE J. Sel. Top. Appl. Earth Observations Remote Sensing 11 (2018) 3312–3324. https://doi.org/10.1109/JSTARS.2018.2856741.

[20] D. Xiaohui, L. Huapeng, L. Yong, Y. Ji, Z. Shuqing, Comparison of swarm intelligence algorithms for optimized band selection of hyperspectral remote sensing image, Open Geosciences 12 (2020) 425–442. https://doi.org/10.1515/geo-2020-0155.

[21] Y. Wang, L. Wang, H. Xie, C.-I. Chang, Fusion of Various Band Selection Methods for Hyperspectral Imagery, Remote Sensing 11 (2019) 2125. https://doi.org/10.3390/rs11182125.

[22] Q. Wang, F. Zhang, X. Li, Optimal Clustering Framework for Hyperspectral Band Selection, IEEE Trans. Geosci. Remote Sensing (2018) 1–13. https://doi.org/10.1109/TGRS.2018.2828161.

[23] R.M. Torres, P.W. Yuen, C. Yuan, J. Piper, C. McCullough, P. Godfree, Spatial Spectral Band Selection for Enhanced Hyperspectral Remote Sensing Classification Applications, J. Imaging 6 (2020) 87. https://doi.org/10.3390/jimaging6090087.

[24] W. Sun, Q. Du, Hyperspectral Band Selection: A Review, IEEE Geosci. Remote Sens. Mag. 7 (2019) 118–139. https://doi.org/10.1109/MGRS.2019.2911100.

[25] A.-K. Mahlein, T. Rumpf, P. Welke, H.-W. Dehne, L. Plümer, U. Steiner, E.-C. Oerke, Development of spectral indices for detecting and identifying plant diseases, Remote Sensing of Environment 128 (2013) 21–30. https://doi.org/10.1016/j.rse.2012.09.019.

[26] R. Meng, Z. Lv, J. Yan, G. Chen, F. Zhao, L. Zeng, B. Xu, Development of Spectral Disease Indices for Southern Corn Rust Detection and Severity Classification, Remote Sensing 12 (2020) 3233. https://doi.org/10.3390/rs12193233.

[27] A. Huete, C. Justice, H. Liu, Development of vegetation and soil indices for MODIS-EOS, Remote Sensing of Environment 49 (1994) 224–234. https://doi.org/10.1016/0034-4257(94)90018-3.

[28] J. Rouse, R.H. Haas, D. Deering, J.A. Schell, J. Harlan, Monitoring the Vernal Advancement and Retrogradation (Green Wave Effect) of Natural Vegetation. [Great Plains Corridor] 1973.

[29] F. Boochs, G. Kupfer, K. Dockter, W. Kühbauch, Shape of the red edge as vitality indicator for plants, International Journal of Remote Sensing 11 (1990) 1741–1753. https://doi.org/10.1080/01431169008955127.

[30] Y. Chen, Q. Wang, Y. Wang, S.-B. Duan, M. Xu, Z.-L. Li, A Spectral Signature Shape-Based Algorithm for Landsat Image Classification, IJGI 5 (2016) 154. https://doi.org/10.3390/ijgi5090154.

[31] F.A. Kruse, Spectral-feature-based analysis of reflectance and emission spectral libraries and imaging spectrometer data, in: Algorithms and Technologies for Multispectral, Hyperspectral, and Ultraspectral Imagery XVIII, Baltimore, Maryland, SPIE, 2012, 83901F-83901F-10.

[32] S. Andrés, D. Arvor, I. Mougenot, T. Libourel, L. Durieux, Ontology-based classification of remote sensing images using spectral rules, Computers & Geosciences 102 (2017) 158–166. https://doi.org/10.1016/j.cageo.2017.02.018.

[33] T.M. Berhane, C.R. Lane, Q. Wu, B.C. Autrey, O.A. Anenkhonov, V.V. Chepinoga, H. Liu, Decision-Tree, Rule-Based, and Random Forest Classification of High-Resolution Multispectral Imagery for Wetland Mapping and Inventory, Remote Sensing 10 (2018) 580. https://doi.org/10.3390/rs10040580.

[34] W. Cui, M. Yao, Y. Hao, Z. Wang, X. He, W. Wu, J. Li, H. Zhao, C. Xia, J. Wang, Knowledge and Geo-Object Based Graph Convolutional Network for Remote Sensing Semantic Segmentation, Sensors 21 (2021) 3848. https://doi.org/10.3390/s21113848.

[35] G. Ghazaryan, O. Dubovyk, F. Löw, M. Lavreniuk, A. Kolotii, J. Schellberg, N. Kussul, A rule-based approach for crop identification using multi-temporal and multi-sensor phenological metrics, European Journal of Remote Sensing 51 (2018) 511–524. https://doi.org/10.1080/22797254.2018.1455540.

[36] P.F. Houhoulis, W. Michener, Detecting wetland change: A rule-based approach using NWI and SPOT-XS data, Photogrammetric Engineering and Remote Sensing 66 (2000) 205–211.

[37] S. Liu, Q. Shi, Multitask Deep Learning With Spectral Knowledge for Hyperspectral Image Classification, IEEE Geosci. Remote Sensing Lett. 17 (2020) 2110–2114. https://doi.org/10.1109/LGRS.2019.2962768.

[38] R.N. Clark, G.A. Swayze, K.E. Livo, R.F. Kokaly, S.J. Sutley, J.B. Dalton, R.R. McDougal, C.A. Gent, Imaging spectroscopy: Earth and planetary remote sensing with the USGS Tetracorder and expert systems, J.-Geophys.-Res. 108 (2003). https://doi.org/10.1029/2002JE001847.

[39] R.N. Clark, T.L. Roush, Reflectance spectroscopy: Quantitative analysis techniques for remote sensing applications, J.-Geophys.-Res. 89 (1984) 6329–6340. https://doi.org/10.1029/JB089iB07p06329.

[40] P.-N. Tan, M. Steinbach, A. Karpatne, V. Kumar, Introduction to data mining, Second edition, Pearson, NY NY, 2019.

[41] J.-J. Ponciano, M. Roetner, A. Reiterer, F. Boochs, Object Semantic Segmentation in Point Clouds—Comparison of a Deep Learning and a Knowledge-Based Method, IJGI 10 (2021) 256. https://doi.org/10.3390/ijgi10040256.

[42] R. Dvorak, E. Kosior, L. Moody, Development of NIR Detectable Black Plastic Packaging, 2011. http://www.wrap.org.uk/sites/files/wrap/Recyclability_of_Black_Plastic_Summary.pdf.

[43] O. Rozenstein, E. Puckrin, J. Adamowski, Development of a new approach based on midwave infrared spectroscopy for post-consumer black plastic waste sorting in the recycling industry, Waste Manag. 68 (2017) 38–44. https://doi.org/10.1016/j.wasman.2017.07.023.

[44] S. Zhu, H. Chen, M. Wang, X. Guo, Y. Lei, G. Jin, Plastic solid waste identification system based on near infrared spectroscopy in combination with support vector machine, Advanced Industrial and Engineering Polymer Research 2 (2019) 77–81. https://doi.org/10.1016/j.aiepr.2019.04.001.

[45] X. Wu, J. Li, L. Yao, Z. Xu, Auto-sorting commonly recovered plastics from waste household appliances and electronics using near-infrared spectroscopy, Journal of Cleaner Production 246 (2020) 118732. https://doi.org/10.1016/j.jclepro.2019.118732.

[46] N.Y. Kiang, J. Siefert, Govindjee, R.E. Blankenship, Spectral signatures of photosynthesis. I. Review of Earth organisms, Astrobiology 7 (2007) 222–251. https://doi.org/10.1089/ast.2006.0105.

[47] D. de Carvalho Lopes, A.J. Steidle Neto, Classification and Authentication of Plants by Chemometric Analysis of Spectral Data, in: Vibrational Spectroscopy for Plant Varieties and Cultivars Characterization, Elsevier, 2018, pp. 105–125.

[48] T. Beghin, J.S. Cope, P. Remagnino, S. Barman, Shape and Texture Based Plant Leaf Classification, in: J. Blanc-Talon, D. Bone, W. Philips, D. Popescu, P. Scheunders (Eds.), Advanced Concepts for Intelligent Vision Systems, Springer Berlin Heidelberg, Berlin, Heidelberg, 2010, pp. 345–353.

[49] P. Salve, P. Yannawar, M. Sardesai, Multimodal plant recognition through hybrid feature fusion technique using imaging and non-imaging hyper-spectral data, Journal of King Saud University - Computer and Information Sciences (2018). https://doi.org/10.1016/j.jksuci.2018.09.018.

[50] A.C. Siravenha, S.R. Carvalho, Plant Classification from Leaf Textures, in: 2016 International Conference on Digital Image Computing: Techniques and Applications (DICTA), Gold Coast, Australia, IEEE, 112016, pp. 1–8.

[51] Q. Wang, F. Zhang, X. Li, Hyperspectral Band Selection via Optimal Neighborhood Reconstruction, IEEE Trans. Geosci. Remote Sensing 58 (2020) 8465–8476. https://doi.org/10.1109/TGRS.2020.2987955.